\documentclass[conference]{IEEEtran}
\usepackage{amsmath,amssymb}
\usepackage{graphicx}
\usepackage{algorithm}
\usepackage{algpseudocode}
\usepackage{booktabs}
\usepackage{makecell}
\usepackage{enumerate}
\usepackage{subcaption}
\usepackage{cite}
\usepackage{float}
\usepackage{dblfloatfix}

\usepackage[absolute,overlay]{textpos}
\newlength{\figwidth}
\IEEEoverridecommandlockouts

\begin{document}
\title{Time-to-Collision Based Dynamic Obstacle Avoidance Using Pretrained Vision Models for Robots in Unstructured Environments}

\author{
\IEEEauthorblockN{
Erik Jagnandan\textsuperscript{1,*},
Mulugeta Haile\textsuperscript{2,\dag},
Gregory Barber\textsuperscript{2}
Pratik Chaudhari\textsuperscript{1}
}
\IEEEauthorblockA{
\textsuperscript{1}GRASP Laboratory, University of Pennsylvania, PA
}
\IEEEauthorblockA{
\textsuperscript{2}U.S. Army Research Laboratory, Aberdeen Proving Ground, MD
}
\thanks{\textsuperscript{\dag}Corresponding author: mulugeta.a.haile.civ@army.mil} \thanks{\textsuperscript{*}erik.jagnandan@gmail.com}

}

\maketitle


\begin{abstract}

Dynamic obstacle avoidance in unstructured outdoor environments remains a critical challenge for autonomous mobile robots, particularly when large-scale robot-specific training data and simulation-based policies are impractical. We present a data-efficient, interpretable method for vision-based dynamic obstacle avoidance that operates entirely on real-world data, avoiding the sim-to-real transfer problem inherent in simulation-trained policies. Our approach leverages UniDepth, a large pretrained monocular depth estimation model, to produce dense depth maps from RGB video without requiring stereo cameras or LiDAR at inference time. Dynamic obstacle avoidance is achieved by extending the SuperPoint and SuperGlue feature correspondence pipeline to track keypoints across long frame sequences, projecting their 2D pixel-space positions into 3D using camera intrinsics and predicted depth, running bundle adjustment initialized from these 3D keypoints, and computing per-keypoint time-to-collision (TTC). A 2D motion primitive in the ground plane is then selected to move the robot away from the closest point of approach of the minimum-TTC keypoint. Evaluated on real-world data from the M3ED dataset, our pipeline achieves a precision of 0.49 and a recall of 0.38 in identifying frames with a ground truth TTC below 1 second, and correctly generates the evasive motion direction in 84\% of true positive detections. Crucially, it detects at least one frame with TTC less than 1 second for 20 out of 22 unique physical obstacles present in our test sequences. Unlike end-to-end learned methods that demand thousands of hours of robot-specific training data, our approach eliminates model training entirely, requiring only 74 seconds of data for hyperparameter tuning. This demonstrates exceptional data efficiency while preserving interpretable and generalizable behavior across diverse obstacle types.

\end{abstract}

\section{Introduction}

Transformers have emerged as powerful architectures for robotic perception and control, achieving state-of-the-art performance and strong generalization across tasks. Prior work in robotic manipulation and visual navigation shows that transformer-based models can outperform traditional convolutional neural network (CNN) and recurrent neural network (RNN) based approaches when trained on sufficient data \cite{brohan2023rt1roboticstransformerrealworld} \cite{shah2023vintfoundationmodelvisual} \cite{bonatti2022pactperceptionactioncausaltransformer} \cite{xiao2022maskedvisualpretrainingmotor} \cite{bucker2022lattelanguagetrajectorytransformer}. However, collecting the large-scale robot-specific datasets required to train such models from scratch is prohibitively expensive. For example, Robot Transformer \cite{brohan2023rt1roboticstransformerrealworld} required 17 months of continuous data collection from 13 robots, creating a major bottleneck to scalability.


A common alternative is to train reinforcement learning (RL) policies in simulation and deploy them in the real world. While simulation removes data constraints, it introduces a “sim-to-real” gap, where performance does not fully transfer due to discrepancies in visual realism, contact dynamics, and environmental diversity between the simulated and real environments. Although techniques such as domain randomization \cite{tobin2017domainrandomization} and SimOpt \cite{yu2019simopt} mitigate this issue, simulation-based training necessarily presents two other key limitations. First, policies learn only from simulated experience, making their behavior unpredictable in real-world scenarios, especially in complex, unstructured environments with unforeseen obstacles. Second, RL policies are typically black-box models with limited interpretability, making it difficult to understand or guarantee their behavior. Together, these challenges hinder robustness and reliability in the real-world deployment of policies trained via RL.

We propose leveraging pretrained vision foundation models to bypass the sim-to-real gap while enabling robust and interpretable performance without large-scale robotic data collection. Specifically, we develop a visual dynamic obstacle avoidance approach using UniDepth \cite{piccinelli2024unidepthuniversalmonocularmetric}, a transformer-based monocular depth model which is trained for geometrically consistent metric depth estimation across domains and camera configurations. We use UniDepth to produce dense depth maps from RGB video, and use SuperPoint \cite{detone2018superpointselfsupervisedpointdetection} and SuperGlue \cite{sarlin2020supergluelearningfeaturematching} to establish keypoint correspondences across frames. By combining these correspondences with camera intrinsics and dense depth from UniDepth, and then running bundle adjustment on the resulting 3D keypoints using the XM solver \cite{han2025buildingromeconvexoptimization}, we track 3D keypoint positions and compute time-to-collision (TTC) for each keypoint. We then propose a 2D motion primitive moving the robot away from the keypoint with the lowest TTC, enabling robust, interpretable dynamic obstacle avoidance. However, while pretrained monocular depth models such as UniDepth offer strong cross-domain generalization, they remain subject to inherent limitations of monocular depth estimation, and our pipeline inherits these issues. In particular, monocular depth performance can degrade in low-texture regions, under extreme lighting variation, or at object boundaries where small pixel-level misalignments can lead to depth discontinuities. Moreover, as UniDepth infers metric scale from learned priors rather than direct geometric triangulation, residual scale bias and local depth inaccuracies can propagate into downstream motion and TTC estimation.

By relying on models pretrained on real-world data, rather than training from scratch in simulators with visuals and contact dynamics of limited realism, our approach avoids the sim-to-real gap associated with training RL policies in simulation. Furthermore, by explicitly estimating 3D structure and motion, it provides interpretable and generalizable behavior across diverse obstacle types, unlike simulation-trained policies that depend on task-specific priors. 

\begin{figure}[t]
    \centering
    \includegraphics[width=0.3\textwidth]{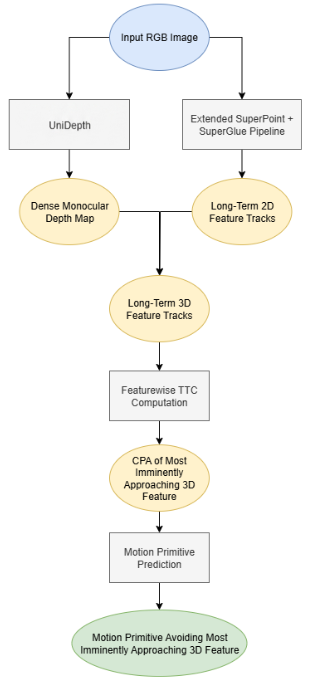}
    \caption{Diagram of the system architecture for our approach.}
    \label{fig:visual_abstract}
\end{figure}

To evaluate generalization and robustness, we target unstructured outdoor environments, and validate our approach on real-world data from the M3ED dataset \cite{Chaney_2023_CVPR} featuring forests and urban scenes.


Notably, dynamic obstacle avoidance can be addressed using sensor modalities other than vision, such as stereo depth or 3D LiDAR. However, we focus on vision due to its advantages in range and density. Monocular depth models can provide useful estimates up to 20–30 meters, whereas stereo depth becomes unreliable beyond 5–10 meters, limiting reaction time to incoming obstacles. Additionally, while LiDAR provides accurate depth for large objects, its sparse measurements may miss smaller or distant obstacles. In contrast, RGB-based depth estimation yields dense scene representations, improving detection of small or far-away obstacles—critical in unstructured environments.

Our contributions are as follows:

\begin{enumerate}[1]
  \item We demonstrate that pretrained vision foundation models enable interpretable and generalizable vision-based robotic control without sim-to-real gaps or large-scale data collection.
  \item We develop a novel method for TTC-based visual dynamic obstacle avoidance built using monocular depth estimation (UniDepth), bundle adjustment (XM Solver), and keypoint correspondence (SuperPoint and SuperGlue).
  \item We validate our approach on data from complex, unstructured, outdoor environments, demonstrating the ability to detect obstacles approaching the camera and to generate corrective actions moving the robot away from these obstacles.
\end{enumerate}

\section{Related Work}

\subsection{Large, Robotics-Specific Models}

Transformer-based models have become central to robotic learning. Robot Transformer \cite{brohan2023rt1roboticstransformerrealworld} and Visual Navigation Transformer (ViNT) \cite{shah2023vintfoundationmodelvisual} are prominent architectures trained end-to-end for robotic manipulation and visual navigation, respectively. Robot Transformer maps visual observations and natural language instructions to low-level actions, while ViNT learns goal-conditioned navigation from RGB inputs. Both demonstrate strong generalization across tasks and environments, but rely on large-scale robot-specific data collection, which poses a significant barrier to their implementation. For example, the data used to train Robot Transformer (collected over 17 months using 13 robots) involved $\sim$130k real-world demonstrations across 700 different tasks.

In contrast to these fully end-to-end systems, Perception-Action Causal Transformer (PACT) \cite{bonatti2022pactperceptionactioncausaltransformer} improves efficiency through a modular design that separates representation learning from task execution. A transformer encoder is pretrained on predicting future robot states and actions, and the features produced by this encoder are passed to small multilayer perceptrons (MLPs), each of which performs a specific downstream task (localization, mapping, or navigation). This modular approach in which all tasks can be performed while only requiring one large transformer encoder allows PACT to be computationally efficient while maintaining strong performance. However, while each MLP requires only a limited amount of training data due to their small, task-specific nature, the PACT approach still depends on a large, robotics-specific dataset to train its encoder.


\subsection{Applying Vision and Language Models to Robotic Tasks}


Recent work explores using pretrained vision and language models as general-purpose backbones for robotics. For example, Masked Visual Pretraining for Motor Control \cite{xiao2022maskedvisualpretrainingmotor} trains a vision transformer via masked image reconstruction, then freezes the encoder and learns RL policies on top of it. This approach achieved stronger performance and generalization than supervised approaches, while also not requiring any labels or task-specific fine-tuning of the encoder.

Similarly, LAnguage Trajectory TransformEr (LATTE) \cite{bucker2022lattelanguagetrajectorytransformer} integrates pretrained models such as BERT \cite{devlin2019bertpretrainingdeepbidirectional} and CLIP \cite{radford2021learningtransferablevisualmodels} to map language instructions and visual inputs to robot trajectories. By keeping pretrained components frozen and training only task-specific modules, LATTE demonstrates strong cross-task and cross-platform generalization.

These works avoid the limitation of scarce robot-specific data by leveraging internet-scale pretrained models from other modalities. We similarly leverage a pretrained vision model, in our case UniDepth, but instead to enable dynamic obstacle avoidance.


\subsection{Reinforcement Learning for Robot Control}


Simulators such as MuJoCo \cite{todorov2012mujoco}, CARLA \cite{dosovitskiy2017carla}, and IsaacGym \cite{macklin2020isaacgym} enable fast and scalable generation of training data across diverse robotic tasks. Many model-free RL algorithms, including DDPG \cite{lillicrap2015ddpg}, PPO \cite{schulman2017ppo}, and SAC \cite{haarnoja2018sac}, have demonstrated strong performance in these environments. However, their reliance on precise and often idealized simulations presents a challenge when transitioning to the real world.

To mitigate the sim-to-real gap, domain randomization techniques such as those proposed by Tobin et al. \cite{tobin2017domainrandomization} introduce variability in textures, lighting, and physics parameters during training to improve generalization. SimOpt \cite{yu2019simopt} further refines this idea by adapting simulation parameters based on real-world observations using Bayesian optimization, allowing for better alignment between simulated and real-world dynamics.

Other works explore sim-to-real transfer by learning representations that are robust across domains. For instance, RAD \cite{laskin2020rad} applies image augmentations to improve sample efficiency and generalization in visual RL, while DreamerV3 \cite{hafner2023dreamerv3} uses a world model to plan and act using latent representations, demonstrating high data efficiency in simulation.

Despite these advances, real-world deployment often requires extensive tuning and system-specific engineering. These approaches, while effective in simulation, typically provide limited guarantees of robustness or interpretability once transferred to the real world. As such, we consider the alternative approach of leveraging models pretrained on real-world data distributions to bypass the challenges of simulation altogether.

\subsection{Time-to-Collision-Based Obstacle Avoidance}

Time-to-collision (TTC) estimation serves as a robust, scale-invariant metric for reactive robotic navigation and obstacle avoidance. Fundamentally, TTC defines the remaining time before an observer collides with an object under the assumption of constant relative velocity. Historically grounded in biological vision \cite{lee1976theory}, modern approaches to TTC in robotics generally bifurcate into geometric frameworks and learning-based paradigms.

Geometric approaches rely on explicit spatial and temporal measurements to compute TTC. Rather than computing absolute distance or velocity independently, these approaches determine TTC as the ratio of an object's visual scale to its rate of expansion over time. Classical monocular methods leverage optical flow divergence or image brightness gradients to deduce these expansion rates without explicit depth estimation \cite{horn2007direct}. However, pure visual estimation degrades during agile robotic maneuvers, as camera rotations (e.g., pitching or yawing) generate pixel motion that mimics forward expansion. To address this, contemporary geometric frameworks utilize sensor fusion, combining visual inputs with inertial data, stereo vision, or event-based sensors, to isolate and remove rotational motion artifacts, ensuring TTC is computed strictly from actual forward translation toward an obstacle \cite{byrne2009expansion, clady2014asynchronous}. While these methods offer deterministic interpretability and metric consistency, they often require careful sensor calibration and can be computationally expensive on resource-constrained platforms.

Conversely, learning-based approaches bypass explicit geometric tracking by training neural networks to predict TTC maps or binary collision boundaries directly from visual sequences \cite{yang2020up, badki2021binary}. These networks implicitly capture scene semantics and complex motion cues from monocular inputs, eliminating the need for specialized hardware. However, these methods are often constrained by the diversity of their training datasets, can exhibit brittle generalization in novel environments, and lack the transparent interpretability required for safety-critical robotic deployment.

Our approach bridges these two paradigms by combining the generalization strengths of learning-based models with the explicit interpretability of geometric frameworks. Rather than deploying an end-to-end neural network to directly predict abstract TTC maps or collision boundaries, we leverage UniDepth to extract rich geometric information in the form of metric monocular depth maps. From these estimated depths, we compute TTC through direct geometric reasoning as the ratio of an keypoint's metric depth to its relative velocity along the camera's optical axis. By decoupling depth estimation from the TTC calculation, our method retains the deterministic interpretability of geometric approaches while capitalizing on the robust, zero-shot generalization capabilities of modern monocular depth estimators.

\section{Preliminaries}

\subsection{SuperPoint and SuperGlue}

SuperPoint \cite{detone2018superpointselfsupervisedpointdetection} is a self-supervised keypoint detector and descriptor that replaces handcrafted methods (e.g., ORB \cite{6126544} and SIFT \cite{inbook}) with a fully convolutional network.  It is trained without labeled keypoints by using Homographic Adaptation, where multiple homographic transformations of an image are used to generate pseudo-ground-truth keypoints, enabling robust and repeatable feature detection under viewpoint and illumination changes.

SuperGlue \cite{sarlin2020supergluelearningfeaturematching} extends this pipeline with a transformer-based graph neural network for context-aware feature matching. By jointly reasoning over all keypoints in two images using attention, it achieves strong correspondence results, including in ambiguous or low-texture regions. Together, SuperPoint and SuperGlue form a widely used pipeline for feature matching in SLAM, structure-from-motion, and visual odometry.


\subsection{UniDepth}

UniDepth \cite{piccinelli2024unidepthuniversalmonocularmetric} is a monocular metric depth estimation model designed to produce geometrically consistent depth predictions across diverse camera intrinsics and environments. Unlike approaches that primarily focus on relative depth and require post-hoc scaling, UniDepth is trained to directly predict metric depth by conditioning on camera parameters, allowing it to generalize across varying focal lengths and fields of view. Architecturally, UniDepth leverages a transformer-based vision backbone combined with a lightweight decoder that regresses dense metric depth maps while explicitly incorporating intrinsic calibration information into the prediction process.

A central contribution of UniDepth is its large-scale, geometry-consistent training strategy. The model is trained on a mixture of real and synthetic datasets with accurate metric supervision, spanning a wide range of scenes, viewpoints, and camera configurations. By enforcing consistency across camera intrinsics and scene scales, UniDepth learns a unified depth representation that transfers effectively across domains. The training pipeline also emphasizes scale-aware augmentations and normalization schemes that prevent overfitting to specific camera setups, improving robustness to distribution shifts.

In contrast to methods that rely heavily on pseudolabels derived from sparse LiDAR or structure-from-motion pipelines, UniDepth focuses on preserving metric fidelity during training and explicitly modeling geometric constraints. This design enables strong zero-shot generalization to unseen datasets and camera configurations, making UniDepth particularly attractive for robotic applications such as ours where accurate metric scale and generalization to complex environments are essential.


\subsection{XM Solver}

The XM solver \cite{han2025buildingromeconvexoptimization} is a certifiably optimal optimization engine designed for the Scaled Bundle Adjustment (SBA) problem. While traditional pipelines rely on local gradient-based techniques that frequently get trapped in suboptimal local minima, the XM solver reformulates SBA as a Semidefinite Program (SDP) relaxation. By leveraging a low-rank Riemannian optimization framework based on the Burer-Monteiro factorization and a custom CUDA-accelerated trust-region optimizer, it guarantees a mathematically certifiable global optimum at extreme scale.

In our pipeline, the XM solver serves as the geometric back-end, optimizing the 3D keypoint trajectories provided to it based on UniDepth and SuperPoint+SuperGlue. However, because it operates strictly as a geometric optimizer, its physical accuracy is fundamentally bounded by the quality of its inputs. While the solver guarantees a mathematically optimal solution relative to its objective function, any systematic errors, scale drift, or geometric inconsistencies inherited from front-end models—such as UniDepth—will propagate directly into the final optimized state.

\section{Method}

\subsection{Dataset}

Our work utilizes the Multi-Robot, Multi-Sensor, Multi-Environment Event Dataset (M3ED) introduced by Chaney et al. \cite{Chaney_2023_CVPR}, a dataset designed for event-based perception across diverse environments and robotic platforms.

We select M3ED for its high-fidelity, synchronized multi-sensor data and coverage of real-world ground locomotion in complex terrains. We focus on sequences prefixed with \textit{spot-forest} and \textit{spot-outdoor-day}, which capture a Boston Dynamics Spot quadruped navigating forest environments and diverse urban scenes. These scenarios align closely with our goal of enabling robust perception in unstructured outdoor conditions. Among these data sequences, we use the \textit{spot-forest-easy-1} sequence for tuning the hyperparameters of our approach, and use the remaining 4 \textit{spot-forest} sequences and all 8 \textit{spot-outdoor-day} sequences for evaluation.

\begin{figure}[h]
    \centering
    \includegraphics[width=0.48\textwidth]{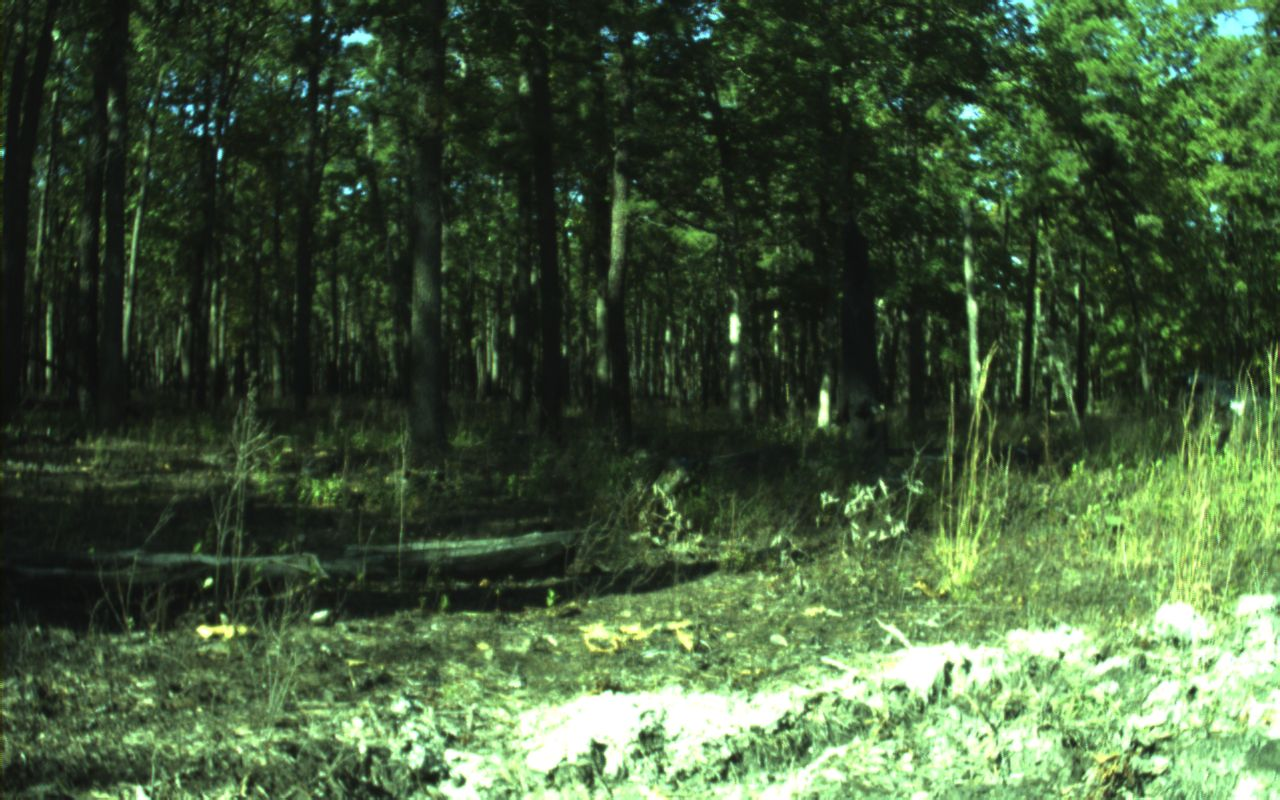}
    \caption{RGB image for frame 0 of the \texttt{spot-forest-easy-1} sequence of the M3ED dataset.}
    \label{fig:rgb_sequence}
\end{figure}

\begin{figure}[h]
    \centering
    \includegraphics[width=0.48\textwidth]{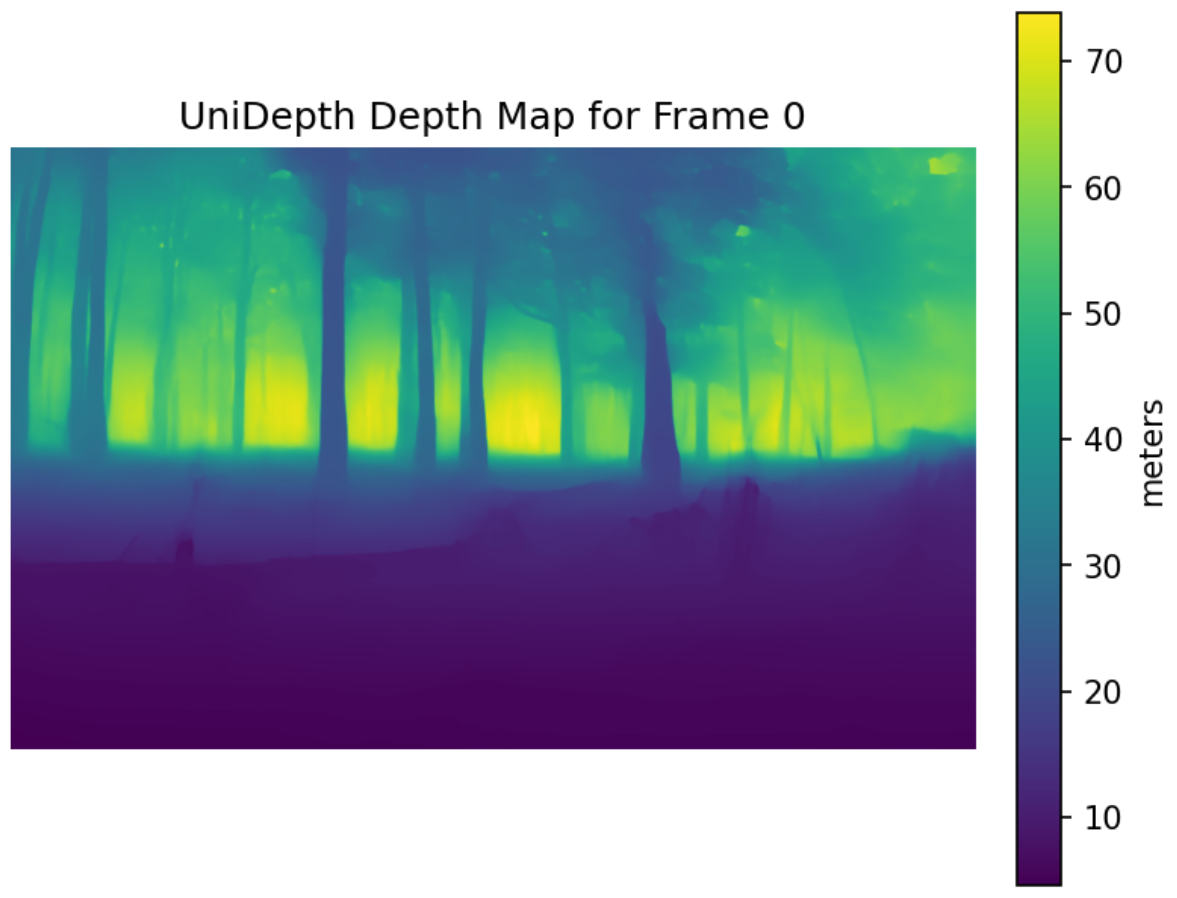}
    \caption{Depth map predicted by UniDepth for frame 0 of the \texttt{spot-forest-easy-1} sequence of the M3ED dataset.}
    \label{fig:naive_depth_anything_sequence}
\end{figure}

\begin{figure}[h]
    \centering
    \includegraphics[width=0.48\textwidth]{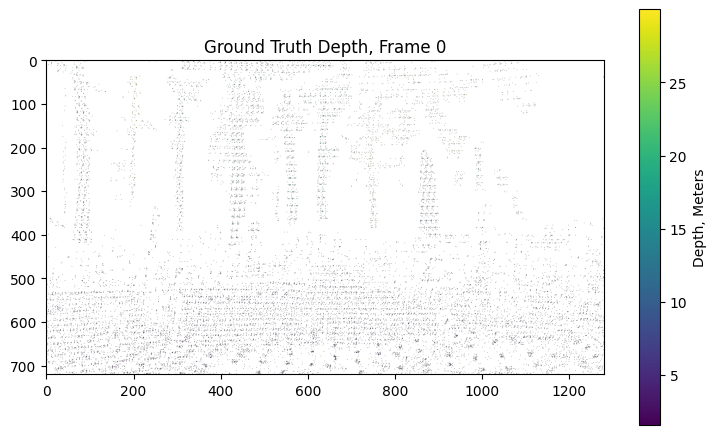}
    \caption{Ground truth depth map for frame 0 of the \texttt{spot-forest-easy-1} sequence of the M3ED dataset. White pixels indicate that there is no ground truth value at that pixel.}
    \label{fig:original_depth_gt_sequence}
\end{figure}

Figures \ref{fig:rgb_sequence}, \ref{fig:naive_depth_anything_sequence}, and \ref{fig:original_depth_gt_sequence} display the RGB image, metric depth map predicted by UniDepth, and the ground truth LiDAR depth for frame 0 of the \textit{spot-forest-easy-1} sequence.


\subsection{TTC-Based Dynamic Obstacle Avoidance}

We initially considered computing the time-to-collision (TTC) for every pixel in the dense depth map at each frame, which would require aligning the content of depth maps across frames to account for robot motion. We experimented with using phase correlation and dense optical flow using RAFT \cite{teed2020raftrecurrentallpairsfield}, but both failed to achieve reliable alignment, making dense TTC estimation infeasible.

\begin{figure*}[!tb]
    \centering

    \begin{subfigure}{0.32\textwidth}
        \includegraphics[width=\linewidth]{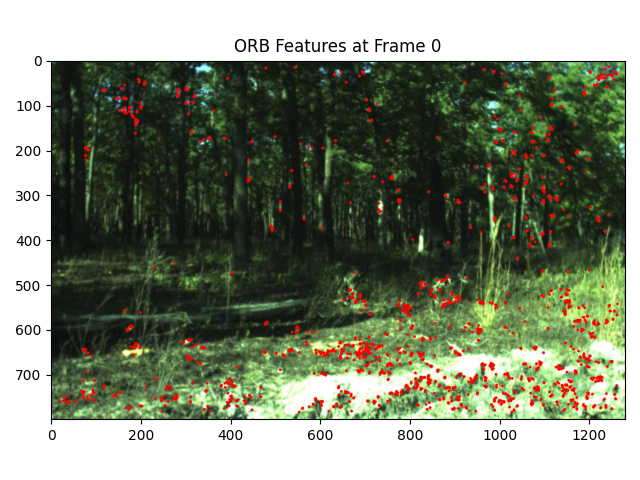}
        \caption*{ORB — Frame 0}
    \end{subfigure}
    \hfill
    \begin{subfigure}{0.32\textwidth}
        \includegraphics[width=\linewidth]{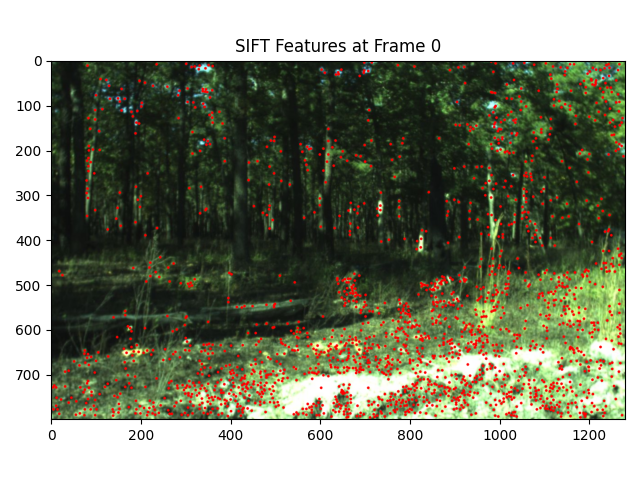}
        \caption*{SIFT — Frame 0}
    \end{subfigure}
    \hfill
    \begin{subfigure}{0.32\textwidth}
        \includegraphics[width=\linewidth]{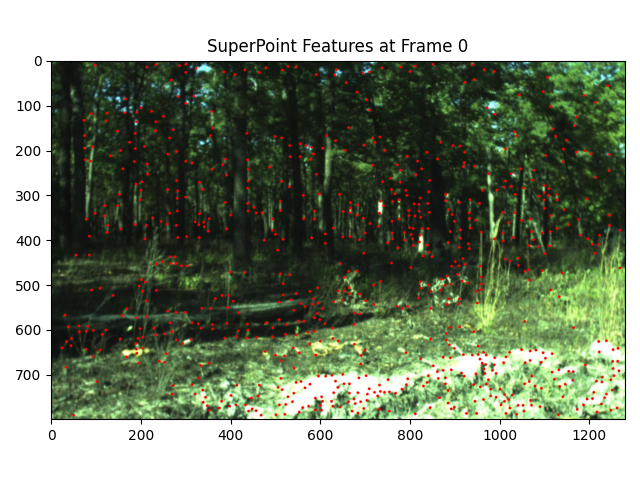}
        \caption*{SuperPoint and SuperGlue — Frame 0}
    \end{subfigure}

    \caption{Keypoints extracted by ORB, SIFT, and SuperPoint for frame 0 of the \texttt{spot-forest-easy-1} sequence of the M3ED dataset. The keypoints shown are those for which correspondences were identified in the next frame. The even distribution of SuperPoint keypoints across the frame, yielding full coverage over all obstacles, motivates our usage of SuperPoint.}
    \label{fig:feature_comparison}
\end{figure*}

We therefore shifted to computing TTC for sparse keypoints within the frame. Handcrafted methods such as ORB \cite{6126544} and SIFT \cite{inbook} exhibited strong bias toward high-brightness, high-contrast regions, failing to capture critical obstacles in dark and low-contrast areas, such as trees. In contrast, SuperPoint and SuperGlue produced keypoints which were evenly distributed both spatially across the frame and across all types of objects in the environment, yielding strong coverage over all obstacles. A comparison of the keypoints extracted by each method is shown in Figure \ref{fig:feature_comparison}.

\begin{figure}[h]
    \centering
    \includegraphics[width=0.48\textwidth]{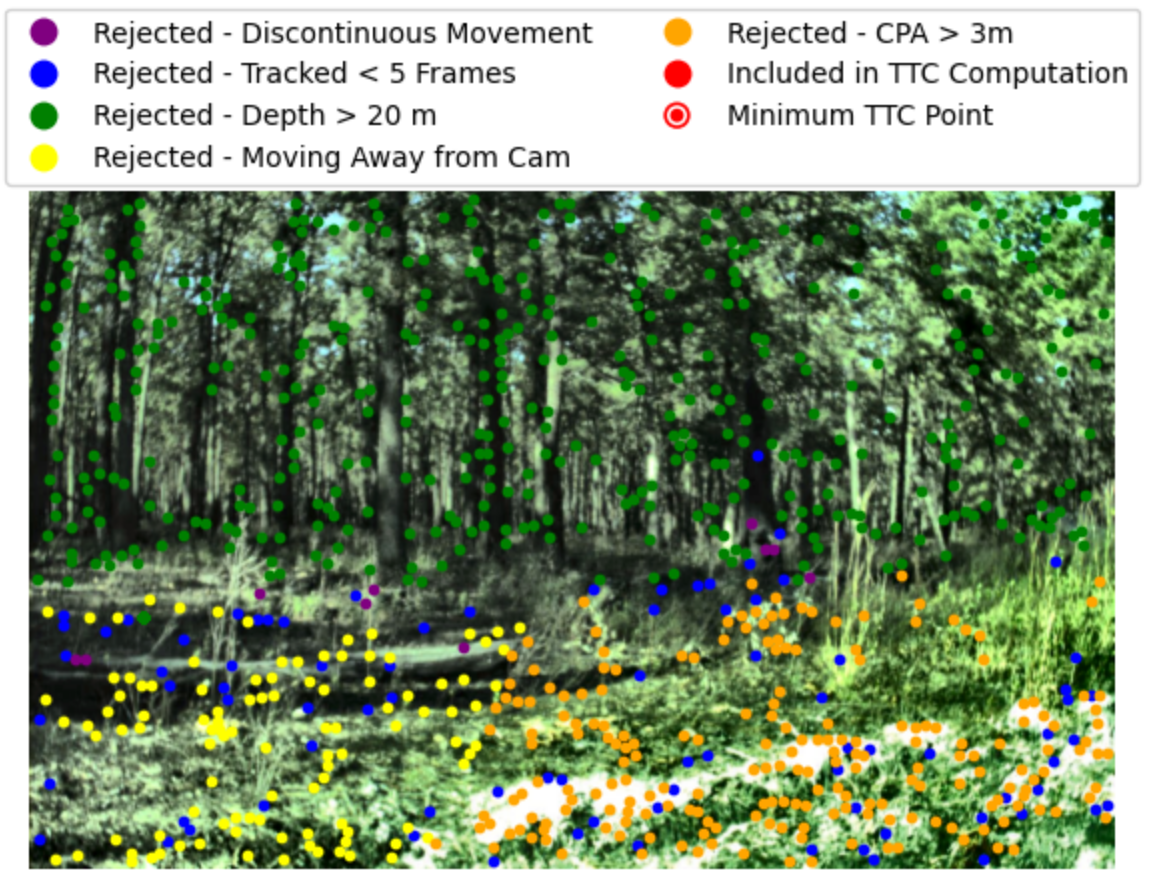}
    \caption{Keypoints for frame 5 of the \texttt{spot-forest-easy-1} sequence of the M3ED dataset. The status of each keypoint is color-coded according to the legend above. Note that not all listed statuses may be present in this specific frame. In this frame, the robot is stationary, so the motion of each point is merely random noise, resulting in no TTC points.}
    \label{fig:interest_points_and_ttc}
\end{figure}

Given the effective performance of SuperPoint and SuperGlue, we extended the pipeline to track keypoints across multiple consecutive frames rather than only between image pairs. This enabled us to maintain temporal correspondences of 2D pixel-space keypoints over time. To lift these tracked 2D pixel coordinates into 3D, we use the predicted metric depth from our fine-tuned UniDepth model together with the known camera intrinsics. Specifically, we assume a standard pinhole camera model:

\begin{equation}
\begin{bmatrix}
u \\
v \\
1
\end{bmatrix}
=
\frac{1}{Z}
\mathbf{K}
\begin{bmatrix}
X \\
Y \\
Z
\end{bmatrix},
\quad
\text{where}
\quad
\mathbf{K}
=
\begin{bmatrix}
f_x & 0 & c_x \\
0 & f_y & c_y \\
0 & 0 & 1
\end{bmatrix}.
\end{equation}

Here, $(u, v)$ are pixel coordinates, $(X, Y, Z)$ are the corresponding 3D camera-frame coordinates, $Z$ is the depth predicted by UniDepth, and $\mathbf{K}$ is the intrinsic calibration matrix with focal lengths $(f_x, f_y)$ and principal point $(c_x, c_y)$. Rearranging this expression yields the back-projection equations used in our implementation:

\begin{equation}
X = \frac{(u - c_x) Z}{f_x}, 
\qquad
Y = \frac{(v - c_y) Z}{f_y}, 
\qquad
Z = Z.
\end{equation}

Using these equations, each tracked 2D keypoint is mapped to a 3D point in the camera coordinate frame, enabling the estimation of its 3D trajectory and velocity over time. To maximize the accuracy of these trajectories prior to TTC estimation, we perform Scaled Bundle Adjustment (SBA) using the XM solver \cite{han2025buildingromeconvexoptimization} over a sliding window of the last five frames. To ensure data reliability, keypoint trajectories tracked for fewer than five frames, as well as points exceeding a depth of 20 meters (which are too distant to represent immediate obstacles requiring evasive action) are excluded from both the optimization and subsequent TTC computation. Furthermore, to mitigate the impact of input noise on the XM solver, we filter out any 3D keypoints exhibiting a frame-to-frame displacement greater than 2 meters. Because such large discontinuities signify inaccuracies stemming from imperfect monocular depth estimation, removing these outliers ensures that the geometric constraints fed into the solver are highly accurate, allowing it to converge to the true physical solution. Finally, whenever these high-displacement anomalies are detected, the associated 3D trajectories are reinitialized to prevent corrupted historical data from impacting future computations.


To perform the TTC computation, we first compute the 3D velocity of the post-bundle-adjustment keypoints via linear regression of their position sequence. For each such keypoint, we compute its closest point of approach (CPA) with the camera and discard keypoints with CPA $>$3 m, as they pose no collision risk. This threshold of 3 meters was selected based on tuning on the \textit{spot-forest-easy-1} sequence which we used for hyperparameter fitting. TTC for each remaining keypoint $i$ is computed as:
\begin{equation}
TTC(i) = \frac{d(i)}{max(-v(i), 0)}
\end{equation}
where $d(i)$ and $v(i)$ are the predicted monocular depth and the velocity computed via linear regression, respectively, at keypoint $i$. This yields finite TTC only for points approaching the camera, and undefined TTC for points which are stationary or moving away, as they will never collide with the camera.

Finally, we select the keypoint with the minimum TTC, and generate a motion primitive (specifically, a 2D motion vector) in the ground plane which moves the robot in the opposite direction to the CPA of this minimum-TTC keypoint. The full system architecture is shown in Figure \ref{fig:visual_abstract}. Example frames depicting the keypoints and their usage in the TTC computation are shown in Figure \ref{fig:interest_points_and_ttc}.


\section{Results}


\subsection{Evaluation Criteria}

We evaluate our approach on all \textit{spot-forest} and \textit{spot-outdoor-day} sequences from the M3ED dataset, except for \textit{spot-forest-easy-1}, which was used for hyperparameter tuning. To enable an objective evaluation of our obstacle avoidance pipeline, we define a time-to-collision (TTC) threshold below which the robot is required to execute an evasive maneuver. The selection of this threshold is specific to the robot platform, as the robot’s speed and kinematic constraints determine how quickly it can react and alter its trajectory once a potential collision is detected. For the Spot robot, which typically operates at approximately 1 m/s (with an official maximum speed of 1.6 m/s), we set the TTC threshold to 1 second. At this nominal operating speed, a 1-second TTC corresponds to a 1-meter physical buffer, providing sufficient distance to initiate and complete an evasive maneuver.

The obstacle avoidance task is therefore formulated as: (i) correctly identifying frames that require evasive action (TTC $<$ 1 s), and (ii) generating an appropriate motion primitive in response. To provide ground truth labels for this evaluation, we use the LiDAR data from M3ED to compute the TTC for all LiDAR points for each frame, select the point with the minimum TTC, and if this value is less than 1 second, define the ground-truth motion primitive as the planar action that moves the robot opposite to the closest point of approach (CPA) of that minimum-TTC point. We evaluate our obstacle avoidance pipeline both on its binary classification accuracy in detecting frames that require evasive action (TTC $<$ 1 s), and, specifically for frames correctly classified as requiring action, the directional consistency between the predicted and LiDAR-derived motion primitives. We consider a prediction correct if the two vectors have a positive dot product.

\subsection{Obstacle Avoidance Performance}

Our obstacle avoidance pipeline exhibits a precision of 0.49 in detecting imminent collisions. It performs effectively in minimizing false positives, as among the 40,104 frames with no ground-truth TTC below 1 s across all sequences, the pipeline produces only 190 false positives (0.47\%). However, our approach detects only 185 of the 482 frames with a ground-truth TTC below 1 s, corresponding to a recall of 0.38. While this frame-level recall is low, it does not critically undermine practical deployment, since for 20 out of 22 distinct physical obstacles that yield a TTC below 1 s at any point in any sequence, the pipeline correctly identifies at least one frame as exhibiting a TTC below 1 second. In a real system, evasive action would be triggered upon the first detection of TTC $<$ 1 s, and subsequent missed detections would be inconsequential once the maneuver has been initiated. Tables \ref{tab:ttc_binary_performance} and \ref{tab:confusion_matrix} summarize the binary classification performance for identifying frames requiring evasive action.

\begin{table*}[t]
\centering
\setlength{\tabcolsep}{3pt}
\begin{tabular}{l*{13}{c}}
\hline
\textbf{Metric} & \textbf{easy-2} & \textbf{hard} & \textbf{road-1} & \textbf{road-3} & \textbf{art museum} & \textbf{penno} & \textbf{skatepark-1} & \textbf{skatepark-2} & \textbf{steps} & \textbf{green loop} & \textbf{bridge-1} & \textbf{bridge-2} & \textbf{Total} \\
\hline
True Positives  & 0 & 49 & 0 & 0 & 0 & 16 & 8 & 20 & 0 & 0 & 52 & 40 & 185 \\
False Positives & 9 & 9 & 1 & 2 & 0 & 31 & 4 & 5 & 0 & 0 & 59 & 70 & 190 \\
True Negatives  & 2915 & 2503 & 4477 & 4992 & 3601 & 2869 & 2293 & 1619 & 3836 & 1597 & 4870 & 4341 & 39913 \\
False Negatives & 0 & 65 & 0 & 0 & 14 & 26 & 0 & 8 & 0 & 0 & 102 & 82 & 297 \\
Precision      & 0.00 & 0.84 & 0.00 & 0.00 & N/A & 0.34 & 0.67 & 0.80 & N/A & N/A & 0.47 & 0.36 & 0.49 \\
Recall         & N/A & 0.43 & N/A & N/A & 0.00 & 0.38 & 1.00 & 0.71 & N/A & N/A & 0.34 & 0.33 & 0.38 \\
\hline
\end{tabular}
\caption{Summary of performance in identifying frames with TTC under 1 second across all sequences. The Total column reports metrics computed from aggregated counts across sequences.}
\label{tab:ttc_binary_performance}
\end{table*}

\begin{table}[h]
\centering
\begin{tabular}{lcc}
\hline
& \textbf{Pred Positive} & \textbf{Pred Negative} \\
\hline
\textbf{Actual Positive} & 38.38\% (185) & 61.62\% (297) \\
\textbf{Actual Negative} & 0.47\% (190) & 99.53\% (39913) \\
\hline
\end{tabular}
\caption{Confusion Matrix for Prediction of Frames with TTC $< 1$ s. Percentages are relative to the sum across each row.}
\label{tab:confusion_matrix}
\end{table}

\begin{figure}[h]
    \centering
    \includegraphics[width=0.48\textwidth]{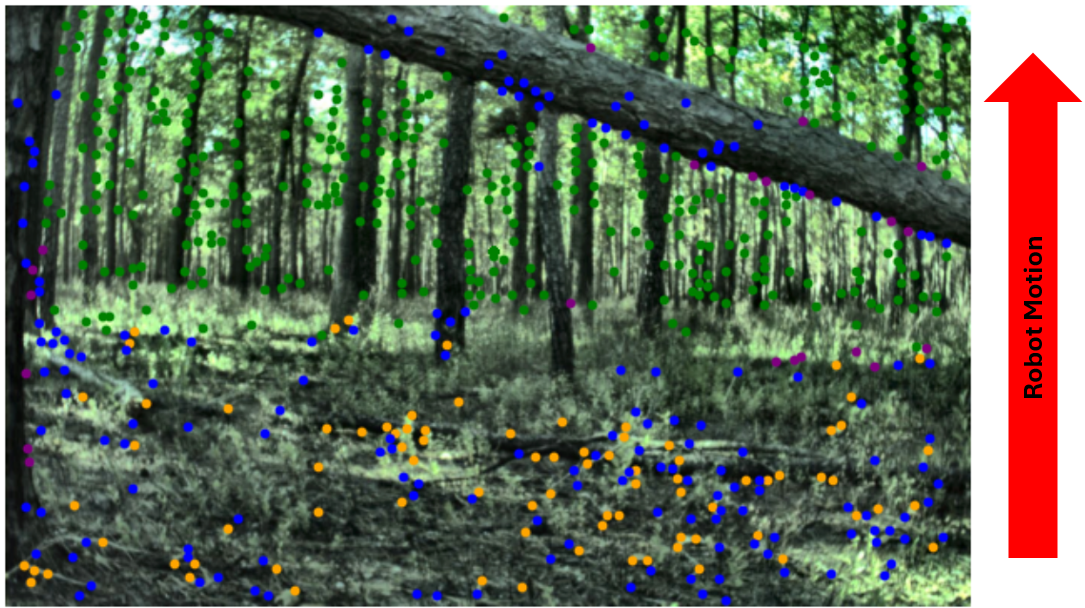}
    \caption{Keypoints for frame 345 of the \texttt{spot-forest-easy-1} sequence of the M3ED dataset. No keypoints on the angled branch were tracked long enough to be included in the TTC computation, leading to no TTC being computed. The keypoints are colored by the same scheme as in Figure \ref{fig:interest_points_and_ttc}.}
    \label{fig:no_features_tracked_long_enough}
\end{figure}

We identify two primary factors contributing to the pipeline’s limited ability to detect frames with TTC below 1 s. First, many missed detections were not caused by inaccurate TTC estimates, but rather by the absence of a TTC estimate altogether. In these cases, no keypoint had been tracked for the required minimum of five consecutive frames, and thus no keypoint qualified for bundle adjustment and the TTC computation. While this was usually a consequence of SuperPoint+SuperGlue failing to maintain long term feature correspondences, it sometimes occurred because inaccurate depth estimates from UniDepth caused discontinuities in the 3D motion of some keypoints, forcing us to reinitialize their trajectories and render them ineligible for bundle adjustment the TTC computation for the next five frames. An example of this failure mode is shown in Figure \ref{fig:no_features_tracked_long_enough}, where no keypoints on the branch had been tracked for five frames, excluding them from the TTC computation. By examining the frame sequence provided in Figure \ref{fig:sequence_frames} in the Appendix, we can observe that this did not occur because of any failure by SuperPoint+SuperGlue, but rather because inaccurate UniDepth depth estimation in frame 344 created a discontinuity in the 3D positions of the keypoints on the branch, causing them to be reinitialized. While this issue could be mitigated by reducing the tracking window below 5 frames, we observed while tuning the tracking window on the \textit{spot-forest-easy-1} sequence that doing so introduced substantial noise from short-lived correspondences, resulting in many false positives while providing minimal reduction in false negatives. 

\begin{figure}[h]
    \centering
    \includegraphics[width=0.48\textwidth]{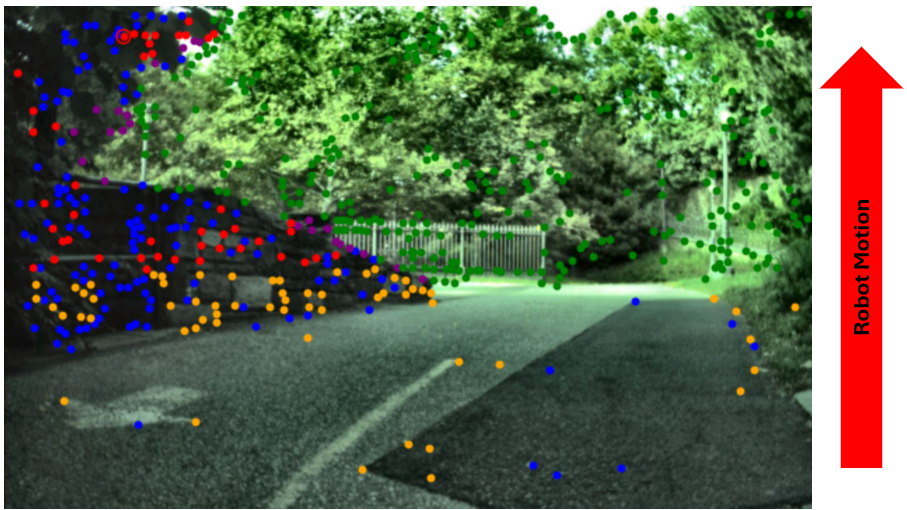}
    \caption{Keypoints for frame 3858 of the \texttt{spot-outdoor-day-under-bridge-2} sequence of the M3ED dataset. Due to inaccurate localization of the circled point on the tree in the top left this frame, a false positive low TTC value is computed for this point. While this inaccuracy comes directly from bundle adjustment, it is likely that is root cause is depth estimation errors made by UniDepth. The keypoints are colored by the same scheme as in Figure \ref{fig:interest_points_and_ttc}.}
    \label{fig:bad_depth_estimation}
\end{figure}

Secondly, in some cases, even correctly tracked keypoints yielded inaccurate TTC values due to imperfections in the optimized 3D keypoint trajectories (see Figure \ref{fig:bad_depth_estimation}), resulting in both false positive and false negative detections. Although filtering steps were implemented to mitigate input noise, systematic inaccuracies or scale inconsistencies inherent to the monocular depth maps from UniDepth still corrupted the underlying data constraints. Because the XM solver operates strictly as a geometric optimizer, it converges to a mathematically optimal solution relative to these flawed constraints, thereby producing physically inaccurate 3D trajectories in challenging scenes. Consequently, the small number of false positive TTC predictions below 1 s generally arose from extreme instances of this depth-induced trajectory distortion.

With respect to selecting the appropriate evasive maneuver once a TTC below 1 s was detected, the pipeline performed strongly, as indicated in Table \ref{tab:evasive_action_performance}. Among the 185 true positive detections, 156 resulted in motion primitives that correctly directed the robot away from the closest point of approach (CPA), as indicated by a positive dot product with the ground-truth motion primitive vector. The failure cases arose from errors in estimating the motion of the approaching keypoint or identifying the wrong keypoint as having the minimum TTC in the frame.

\begin{table*}[t]
\centering
\setlength{\tabcolsep}{3pt}
\begin{tabular}{l*{13}{c}}
\hline
\textbf{Metric} & \textbf{easy-2} & \textbf{hard} & \textbf{road-1} & \textbf{road-3} & \textbf{art museum} & \textbf{penno} & \textbf{skatepark-1} & \textbf{skatepark-2} & \textbf{steps} & \textbf{green loop} & \textbf{bridge-1} & \textbf{bridge-2} & \textbf{Total} \\
\hline
TP TTC $<$ 1s   & 0   & 49   & 0   & 0 & 0   & 16   & 8   & 20 & 0   & 0   & 52   & 40   & 185 \\
Correct Actions & 0   & 42   & 0   & 0 & 0   & 12   & 7   & 18 & 0   & 0   & 45   & 32   & 156 \\
Success Rate   & N/A & 0.86 & N/A & N/A & N/A & 0.75 & 0.88 & 0.90 & N/A & N/A & 0.87 & 0.80 & 0.84 \\
\hline
\end{tabular}
\caption{Summary of performance in predicting correct evasive actions for frames across all sequences which were true positives for TTC under 1 second.}
\label{tab:evasive_action_performance}
\end{table*}

\subsection{Data Efficiency}

A key strength of our approach is its data efficiency. In contrast to methods such as Robot Transformer \cite{brohan2023rt1roboticstransformerrealworld}, which require large, robot-specific datasets for training transformer models from scratch, our pipeline builds upon pretrained models (namely UniDepth, SuperPoint, and SuperGlue), and involves no model training, only requiring hyperparameter tuning on the 74 seconds of data from the \texttt{spot-forest-easy-1} sequence. Other approaches, such as Masked Visual Pretraining for Motor Control \cite{xiao2022maskedvisualpretrainingmotor} avoid the need for real-world robot-specific data by training their RL-based controllers in simulation, but this introduces a sim-to-real gap in terms of photorealism, contact dynamics, and the range of situations encountered in simulation versus real-world deployment.

\section{Conclusion and Future Work}

We have presented a vision-based dynamic obstacle avoidance system for robots operating in complex, unstructured outdoor environments. Our approach combines the UniDepth monocular depth estimator and bundle adjustment via the XM solver with a SuperPoint and SuperGlue feature tracking pipeline to compute per-point time-to-collision estimates and generate corrective motion primitives, all without simulation-based training or large-scale robot-specific data collection.

On the positive side, our pipeline demonstrates several encouraging properties. It successfully detects at least one frame with a threatening TTC for 20 out of 22 unique physical obstacles present in the evaluation sequence, ensuring that almost all obstacles are observed and a corrective action is taken. When a dangerous frame is correctly identified, the pipeline generates the correct evasive motion direction in 84\% of cases, indicating that the underlying TTC and closest point of approach computations are reliable when the depth and tracking conditions are favorable. Additionally, by eliminating model training entirely from the pipeline, our data requirements were restricted solely to the 74-second \texttt{spot-forest-easy-1} sequence used for hyperparameter tuning, underscoring the data efficiency gained by leveraging pretrained vision foundation models.

Nevertheless, we identify several important limitations that must be addressed in future work. The pipeline achieves a recall of only 0.38 in identifying individual frames with a ground truth TTC below one second, primarily due to insufficient long-term feature tracking by SuperPoint and SuperGlue across consecutive frames, outlier depth estimation errors requiring the reinitialization of many 3D keypoints, and inaccurate 3D trajectories as produced by bundle adjustment, ultimately owing to imperfect depth estimates from UniDepth. While we note that a deployed system could initiate evasive action upon the first detected dangerous frame and sustain it thereafter, a recall of 0.38 still represents a meaningful gap, particularly in scenarios where the detection window is narrow or the obstacle approaches rapidly. Improving recall is therefore a primary objective for future iterations of this work. From a system perspective, replacing SuperPoint and SuperGlue with a more robust long-term tracker such as one based on DROID-SLAM \cite{teed2022droidslamdeepvisualslam} features or a Kalman filter-augmented correspondence pipeline would likely improve recall by maintaining feature tracks through challenging low-texture or motion-blurred frames.


Finally, a complete deployment on the Boston Dynamics Spot platform, integrating the motion primitive output with the robot's locomotion controller and measuring end-to-end latency and real-time feasibility, remains as important future work. Such a demonstration would validate the practical utility of the approach and position it as a viable component of a full autonomous navigation stack for legged robots in the wild.

\bibliographystyle{ieeetr}
\bibliography{ref}
\vspace{100pt}

\appendices

\end{document}